# eFedDNN: Ensemble based Federated Deep Neural Networks for Trajectory Mode Inference

Daniel Opoku Mensah[a], Godwin Badu-Marfo[a], Ranwa Al Mallah[b], Bilal Farooq[a]
[a] Laboratory of Innovations in Transportation (LiTrans), Toronto Metropolitan University
Toronto, Canada
[b] Cybersecurtity Lab, Royal Military College
Emails: daniel.mensah@ryerson.ca; gbmarfo@ryerson.ca; ranwa.al-mallah@rmc-cmr.ca;
bilal.farooq@ryerson.ca (corresponding author)

*Abstract*—As the most significant data source in smart mobility systems, GPS trajectories can help identify user travel mode. However, these GPS datasets may contain users' private information (e.g., home location), preventing many users from sharing their private information with a third party. Hence, identifying travel modes while protecting users' privacy is a significant issue. To address this challenge, we use federated learning (FL), a privacy-preserving machine learning technique that aims at collaboratively training a robust global model by accessing users' locally trained models but not their raw data. Specifically, we designed a novel ensemble-based Federated Deep Neural Network (eFedDNN). The ensemble method combines the outputs of the different models learned via FL by the users and shows an accuracy that surpasses comparable models reported in the literature. Extensive experimental studies on a real-world open-access dataset from Montréal demonstrate that the proposed inference model can achieve accurate identification of users' mode of travel without compromising privacy.

## I. INTRODUCTION

In smart mobility systems, there is much concern about privacy when inferring users' mode of travel as planners and engineers regularly collect mobility behaviour information, including mode, activity locations, and trajectories. While collecting massive amounts of information can greatly help in the planning, design, and operations of mobility systems, this approach has the potential to violate users' privacy. For instance, travel survey apps, such as Itinerum [1], usually have cloud computing platforms, where travellers are required to share private information. Even established companies like Tesla, have accidentally leaked drivers' location information, which could be problematic for vehicle owners' privacy [2]. Addressing privacy concerns that arise from transmitting and sharing data containing sensitive information, we incorporate a privacy-preserving machine learning technique known as Federated Learning (FL) [3], where the travellers do not need to expose their raw data for the planning, design, and operations of smart mobility systems.

As an emerging technology, FL allows large-scale nodes such as mobile and edge devices to train and exchange models globally without revealing their local data. FL has been recently adopted in Intelligent Transportation Systems (ITS). Liu et al. [4] proposed a framework known as Federated Gated Recurrent Unit (FedGRU) algorithm for traffic flow prediction using federated learning for privacy preservation. FL was applied by [5] to predict drivers' route choice decisions in route navigation systems, and the feasibility of the FL model was successfully demonstrated by comparing its performance with a centralized server-based model. In the existing literature, research related to mode detection from smartphones has been performed using a variety of approaches. Likewise, Yazdizadeh et al. [6] developed a series of CNN augmented by different ensemble methods to infer travel mode from GPS trajectories gathered by a large-scale smartphone travel survey. However, the authors adopted the centralized ML approach to study various CNN architectures and combined their results using various ensemble model to attain a higher prediction accuracy.

However, the approaches proposed in the literature have some key limitations. First, they prioritize accuracy at the expense of privacy. Secondly, though accuracy is a major priority, different approaches can be achieved with better accuracy, while preserving the privacy of users' data. In order not to violate the European Union's General Data Protection Regulation (GDPR) law on leakage of data, we need to develop new methods to account for the general public growing sense of privacy [4]. To close the research gaps in privacy protection of the existing travel mode inference approaches, while addressing the specificities of travel related trajectory data, we propose a novel ensemble-based federated learning architecture for travel mode inference. This research aims to adopt an ensemble-based structure for FL and utilize the strengths of DNNs to infer users' travel mode using real-world GPS trajectories data obtained residing on the user's smartphone. Specifically, we design a novel ensemble-based Federated Deep Neural Network (eFedDNN) using Long Short Term Memory (LSTM), Gated Recurrent Unit (GRU), and 1 Dimensional Convoluted Neural Network (1D CNN) as the base-learners and Multilayer Perceptron (MLP) as the meta-learner to integrate the optimal global model and capture the spatio-temporal correlation of GPS trajectories data. We conduct an extensive experiment on a real-world open-access dataset to demonstrate the performance of the proposed ensemble-based FL for travel mode inference compared to vanilla FL and non-FL methods.

## II. Background

Several learning paradigms exist for the training of machine learning models [7]. The common approach is the central server paradigm, where the training and inference happen on the server. The second paradigm is where the training happens on the server, but the inference is done on the user's device because the final trained model is sent to all the users. The last paradigm is when training and inference happen on the device. Distributed deep learning consists in large-scale DNNs training where the central server (also known as chief) simply splits the workload amongst the workers at every iteration [8]. After sending portions of the dataset to the workers, the chief then averages the gradients received from the workers and updates the weights of the model in order to then broadcast the new weights to all workers in the next iteration. In this setting, the workers train the model collaboratively, but the data are held by the chief so privacy issues remain. To solve the data access problem, FL was introduced. FL enables the training of a central model without exchanging private local data samples. In FL, a statistical model (e.g. linear regression, neural network, boosting) can be chosen to be trained on the worker nodes. Particularly, in Federated AI, devices collaboratively train Machine Learning (ML) models. This mitigates the data privacy issue and democratizes ML because workers having less data, by collaborating through the FL approach, will be able to train the ML model with workers that hold larger datasets.

Federated learning can be divided into three categories, i.e., horizontal FL, vertical FL and federated transfer learning, based on the distribution characteristics of the data. The vanilla FL setup corresponds to horizontal FL, where data samples of different workers have the same features. Precisely, worker nodes have non-overlapping data points, but they are tracking the same features for the data points. In vertical FL, also called feature-based, workers have the same samples, but features are split among them. Precisely, worker nodes have overlapping data points, but they are tracking different features for those data points. Another FL setting consist of Federated Transfer Learning (FTL), where from the global model, transfer learning is performed to get personalized models [9]. FTL is used when two workers hold datasets that differ not just in terms of samples, but also in terms of feature space. Deep neural networks have been widely adopted in transfer learning to find implicit transfer mechanisms. Model fusion is a particular case of FL. In the most extreme setting, a global neural network needs to be constructed with a single communication round [10]. Model fusion techniques try to establish some correspondence between neurons of different neural networks in order to average them.

Alternatively, ensemble methods [11] are a classic approach for combining predictions of multiple learners. They often perform well in practice even when the ensemble members are of poor quality. Among the well-known ensemble techniques that include boosting and bagging, stacking combine the outputs of the individual learners and lets another algorithm, referred to as the meta-learner, make the final predictions [12]. A super learner is another technique that calculates the final predictions by finding the optimal weights of the base learners by minimizing a loss function based on the cross-validated output of the learners.

The most popular deep learning frameworks and architectures are used to predict and classify time-series or sequential data. In this study, we designed an FL-based ensemble model using LSTM, GRU, and CNN as the base-learners and MLP as the meta-learner. LSTM is a variant of the classic Recurrent Neural Network (RNN) architecture that implements four interactive gates to allow learning sequence labels for extensive time intervals [13]. The unit of this neural network has three gates and a memory cell: input gate, forget gate, output gate, and memory cell. Another modification of RNN that handles time-series data is the Gated Recurrent Unit, proposed by [14]. The GRU has gating units that control the flow of information inside the unit without having separate memory cells. It uses the hidden state to transfer information and only has two gates: a reset and update gate [4]. The last architecture for sequence classification is the One-Dimensional (1D) Convolutional Neural Networks (CNN). While CNN has shown promising success in processing images, it has shown the capability to evaluate patterns in sequences with long-term dependencies. The 1D CNN comprises an input layer, hidden CNN layers, and fully connected layers that end with an output layer.

## III. Methodology

In this work, we assume a dataset of GPS trajectories depicting motion characteristics of members of the population across the main travel modes. First of all, we extract trip characteristics, including "speed, acceleration, and jerk" from sequences of GPS logs recorded from smartphones having GPS capability. The motion characteristics are subsequently segmented into sequence lengths of ten (10) data points having three (3) features (speed, acceleration, jerk). These input features are scaled and normalized to improve training performance in the neural networks. For sequences with less than a length of ten (10), zero padding is performed to achieve a fixed sequences length. The prediction output comprises thirty-five (35) distinct classes of travel modes, as observed in the MTL Trajét dataset from Montréal. These classes are pre-processed and one-hot encoded in the standard format for DNN training. We implement ten (10) worker nodes and a single chief node to aggregate local model updates from the worker nodes. On each worker node, three (3) base-learners comprised of LSTM, CNN 1D, and GRU are deployed. Similarly, these weak learners are deployed at the chief for aggregation. Using the FL approach, the base-learners are trained over twenty (20) iterations. When the chief receives the local updates for each iteration, it aggregates for each of the base-learner models. After aggregation, an ensemble stacking generation is undertaken, which stacks the prediction outputs from the base-learners to a meta-learning framework composed of a multi-layer perceptron. We observe the accuracy of the classification predictions from the aggregated base-learners and the ensemble stacked model when inference is undertaken on the testing dataset

## A. Data Processing

*1) Datasets:* In this work, we employ the real-world open-access GPS trajectory dataset from MTL Trajét 2016 database that has over 33 million locations (primarily GPS) points. The mobility data were collected through a smartphone application involving participants who voluntarily took part in the travel survey. Data on trips having timestamped GPS locations were used in this research. Participants validated their trips by specifying their mode (i.e., walking, biking, car, public transit) and purpose of travel (i.e., school, home, work, business) when trips end. The mode of travel that was completed for trips between these locations was ascertained. Information on the mode of trips (walking, biking, car, public transit) is obtained. Within these trips, the GPS trajectories data obtained are coordinates that are in the form of latitudes and longitudes. This work primarily considered four real travel modes for identification, i.e., walking, biking, car, and public transit. To ensure that training and validation data are not exposed to the evaluation of the proposed FL approaches, users' trip mode data were separated randomly. We first split 5% of all trip data to serve as a proxy dataset to train the global model, and for the rest of the data, 80% is used for training, and 20% is used for testing.

*2) Processing of GPS Records:* The raw GPS data in each user's smartphone consists of ordered points that have been collected over some time. GPS raw data are the user's travel information collected by the smartphone device within a period, i.e., longitude, latitude, and sampling timestamp. Consequently, we alienated the original GPS data with the same travel mode into a trip based on its timestamp. Let G = $G_1, G_2, \ldots, G_n$ represent the GPS record in the segment with length, N. Each GPS record is represented by a triple $G_i$ = ($lat_i$, $long_i$, $t_i$) as the latitude ($lat_i$) and longitude ($long_i$) of the device's location at the time of $t_i$. For two consecutive records $G_i$, $G_{i+1}$, we utilize Vincenty formula [15] to calculate the relative distance:

$D_i$ = Vincenty($lat_i$, $long_i$,; $lat_{i+1}$, $long_{i+1}$) Indicating the time interval between $G_i$ and $G_{i+1}$ as $\Delta t_i$, based on the relative distance ($D_i$), we can calculate the first three motion features based on speed ($S_i$), acceleration ($A_i$), and jerk ($J_i$) of the $R_i$ location using these equations:

$$S_i = \frac{D_i}{\Delta t_i}, \quad 1 \leq i \leq N, S_N = S_{N-1} \quad (1)$$

$$A_i = \frac{S_{i+1} - S_i}{\Delta t_i}, \quad 1 \leq i \leq N, A_N = 0 \quad (2)$$

$$J_i = \frac{A_{i+1} - A_i}{\Delta t_i}, \quad 1 \leq i \leq N, J_N = 0 \quad (3)$$

According to [6], speed is calculated using the distance between each two consecutive GPS points divided by their time interval, whilst acceleration is also defined as the derivative of speed or the rate of change of speed over time. Jerk, or the rate at which acceleration changes, is a crucial component in public transport safety issues such as essential driver maneuvers and passenger balance [16]. In this work, we consider the relative distance, speed, acceleration, and jerk rate. These features constitute the channels of each GPS segment used to train our models.

## B. Federated Learning Training Process

Our study adopted the Federated Averaging Algorithm (FedAvg) for inferring the travel mode of users using GPS datasets of smartphone travel surveys. Using a neural network algorithm, local model parameters are built from each user's (i.e., worker's) own data that is likely to have different numbers of data points (i.e., mode of trips). The FedAvg algorithm combines the local model updates of each worker at a chief node through model averaging. This algorithm is robust to unbalance and non-IID data distributions and can reduce the rounds of communication needed to train a deep network on decentralized data by orders of magnitude. This means that data from a certain worker do not represent the entire population distribution as each worker has a different amount of data. In this research, GPS data from users who are biking are entirely different from users who may be driving.

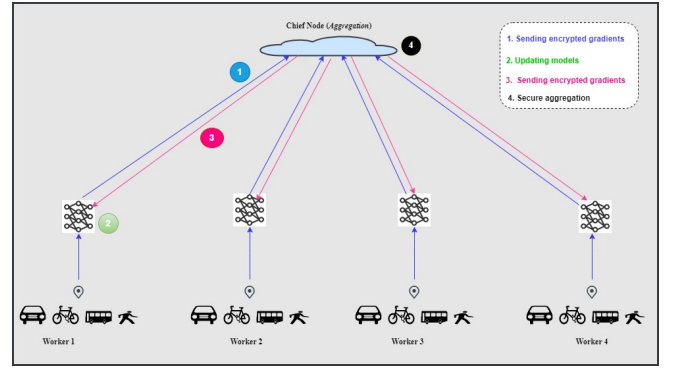

Fig. 1. Federated learning-based travel mode inference architecture

## C. Ensemble Federated Deep Neural Network(eFedDNN) Architecture

Based on the DNN algorithms, we propose an ensemble-based system that integrates the optimal global model and captures the spatio-temporal correlation to improve prediction accuracy. The proposed model is a privacy-preserving FL method known as ensemble Federated Deep Neural Network (eFedDNN). The eFedDNN aims to achieve accurate travel mode inference through FL and DNN without compromising privacy. In the context of this paper, we apply the Gated Recurrent Unit (GRU), Long Short Term Memory (LSTM), the one-dimensional Convolution Neural Network (1D CNN), and the Multi-Layer Perceptron (MLP) deep neural networks. For the travel mode inference problem, we illustrate that each of the DNN models that serves as the local model in the device of each worker in the FL process needs to learn every sample of the GPS dataset located on each smartphone device. This study applies DNNs for the ensemble method because they are nonlinear techniques with enhanced flexibility and have the ability to scale in proportion to the quantity of training data available [17].

The proposed model uses stacked generalization (stacking) to combine the outputs of a set of base-learners and another algorithm, the meta-learner, to make the final model predictions. In stacking, we deploy a meta-learner composed of a multilayer perceptron with multiple neural layers at the chief to accept inputs of the predictions from the average global models of the base learners. The base learners are deployed at both the workers and the chief for aggregation of weights. The meta-learner is only deployed at the chief for ensemble prediction.

Thus, we use three DNNs for the base-model and another DNN for the meta-model. Figure 2 shows a two-tier architecture used in this study with the base-models as GRU, LSTM, and CNN, while MLP is the meta-model. The workers (data owners) have the GPS datasets captured on each device during the FL process. When the different models are trained, the chief uses a synthesized dataset representing the population distribution to proceed with ensemble learning. At the chief node, each neural network classifier predicts inputs sent into the next model for stacking. The process involves computing the unweighted average of the labels from the base-learners and selecting the highest value.

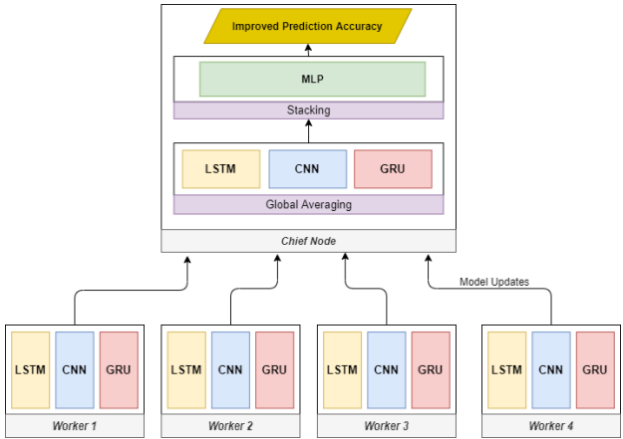

Fig. 2. eFedDNN architecture

### D. Problem Setting

Given $N$ workers $P_1, P_2, \cdots, P_n$ with each of them equipped with a local dataset $D_n$, the problem is to build an ensemble model based on all local datasets with the existence of a *chief*. Let N denote the number of workers where the local dataset of workers is given as $D_n$. Furthermore, each local dataset evolves with the time. Upon a new local dataset arriving, the old dataset will be instantly discarded and no longer available for the training stage. The ensemble Federated Deep Neural Network (eFedDNN) is thus presented in Algorithm 1. The local dataset of worker $n$ is given as $D_n = {x_i, y_{ij}}$. Let $W_k = {w_1, w_2,...,w_k}$ denote the global model of the optimal set $P_k$. As shown in Fig. 2, we use the eFedDNN to find the optimal ensemble model by integrating the global model with the best accuracy after executing the eFedDNN algorithm.

---

**Algorithm 1** ensemble Federated DNN Algorithm
---
**Input:** Worker datasets $\{D_n\}$, the number of iterations and the learning rate, optimizer
**Output:** Global model the ensemble classifier $\{\varnothing\}$;
Randomly divide N worker into batches: B ← $\{B_1, B_2,...,B_n\}$

Initialize $w_o$ (Pre-trained by a public dataset);

**for** $t = 1, 2, 3..., T$ **do**
    // The side of workers
    **for** $n = 1, 2, \cdots, N$ **do**
        Randomly select N worker from D;
        Receive the learner parameters, $w_o$ from the chief;
        Chief broadcasts parameters $w_o$ to workers $B_n$;
        **for** *each worker n in B* **do**
            Accuracy ← validate (output);
        **end**
        Send the updated parameters, $W_k$ to the chief;
    **end**
    // The side of the chief
    Receive updated parameters from all workers;
    Execute each DNN algorithm;
    Obtain the global model set;
    Execute the ensemble learning to find the optimal global model subset;
    Chief sends the new global model to each worker;
**end**
**return** Global model the ensemble classifier $\{\varnothing\}$;

---

The chief initiates the training phase of *eFedDNN*. The chief then builds base learners of specified types after determining the kind of base learners and number of base learners. The learner parameters, $w_o$ are randomly initialized. The parameters are updated iteratively for a total number of rounds via the communications between the chief and all workers. The chief aggregates the local model parameters from all workers of each model, and computes a final model via model averaging. Specifically, at each round t, each worker $P_n$ receives a copy of model parameters from the chief via broadcasting to update its local model parameters. Upon receiving model parameters from all the workers, the chief uses model averaging to compute a final model. To obtain the accuracy of the ensemble model, majority voting is used to predict labels from the base learners and create a final prediction on label with most votes. The chief sends the new global model to each worker.

## IV. RESULTS AND ANALYSIS

In this section, we discuss the results of applying the proposed model to the real-world open-access GPS trajectory dataset. The mobility data are presented and analyzed to test the performance of users' travel mode.

## A. Implementation Details

All simulations that involve data pre-processing and model development are developed in Python programming language using Pytorch for its implementation with GPU support. After the data preparation process, a search is carried out to determine the best network configurations. Dropout layers and their rates, number of nodes (neurons) in each hidden layer, batch size, number of hidden DNN, and dense layers are configured. The models are trained on a Core i7 4GHz CPU and a 16.0 GB memory.

In this experiment, we use 10 workers and train a series of neural networks on each device. To avoid overfitting, we divide the workers into 3 different parts, with one part having 4 workers. Each part, consisting of 3 different workers, is trained with different neural network architecture. We distribute the training data to 10 workers. Each worker has at least one travel mode to simulate the distribution of non-IID data in the real world as much as possible. The proposed eFedDNN algorithm is applied to the real-world open-access GPS trajectory dataset from MTL Trajét 2016 database for performance demonstration. As a federated learning process, the workers (data owners) have the GPS datasets captured on each device. We build ensemble base-learners comprised of three parallel deep learning architectures capable of sequence prediction: LSTM, GRU, and CNN 1D. These base models are tuned to accept input features of travel motion characteristics, namely, speed, acceleration, and jerk estimated from timestamped GPS logs of a mobile travel diary.

Each model consists of two (2) hidden layers with a Rectified Linear Unit (ReLu) activation function. The output layer of the models is designed for multi-modal classification with a linear dense layer having a SoftMax activation. In the eFedDNN algorithm, there exist two optimizers for each DNN model: worker optimizer and chief optimizer. The worker optimizer is used to train the local devices of workers, and the chief optimizer applies averaged workers' updates to the global model on the chief. Adam optimizer was used for both worker and chief with different learning rates: 0.0005 for the worker optimizer, the same rate as the global model, and 0.001 for the chief optimizer. Each worker uses the Adam optimizer to train the proposed model locally for E (E = 10) epochs while the local batch size is set to 30. Since the travel mode inference is a multiclass classification task, categorical cross-entropy is used as the loss function. Federated Averaging (FedAvg) was used to aggregate workers' updates from each device and produce a new global model. Prediction accuracy is measured as the key metric in evaluating model performance.

## B. Interpretation of Results

In this study, we build an ensemble-based FL framework using deep neural networks where LSTM, GRU, and 1D CNN are used as base learners and MLP is used as the meta-learner. To evaluate performance, models are compared based on the prediction accuracy of the test data which is considered the real-world open-access GPS trajectory dataset.

First, we compare the performance of the individual baseline models, i.e., federated learning-based LSTM, GRU, and 1D CNN in identical configurations with each other to determine the model with the best performance. Second, we evaluate the performance of the proposed model, eFedDNN by comparing it in an identical configuration with the individual baseline models. Thus, federated learning-based LSTM, GRU, and 1D CNN are the vanilla DNN models that are used as the baseline models to assess the performance of eFedDNN model. Additionally, we evaluate the performance of the proposed ensemble model with a previous ensemble-based CNN model by [6] on the same data. The individual base models defined are not using any contextual information. However, the GPS trajectories data from the smartphone devices of workers used as the inputs for the FL approach are the same as the one used for the ensemble-based FL. The centralized model of the previous study is applied to the same dataset. Again, we evaluate the performance of the model with centralized model of previous studies that are not based on the same configurations as this study.

*1) Performance Comparison of Proposed Model with Baseline Models:* We developed federated learning-based DNN models as baseline models. Then, we compare the performance of the proposed model with the baseline models of the DNNs. Accuracy, which is the models' prediction power, is shown in Table 1 and Figure 3. It is shown that the proposed ensemble model can infer travel modes more accurately than the individual baseline models. Specifically, the accuracy of the proposed model is 1.0% higher than the vanilla LSTM, which has the highest accuracy among the respective base models.

Additionally, we evaluate the performance of eFedDNN by comparing it with the baseline models using the same number of workers for the training of the model. The proposed ensemble model outperforms the vanilla models, i.e., LSTM, GRU, and 1D CNN, by reducing the error of the mode of travel and further increasing the prediction accuracy. As shown in Figure 3, with the test data, the ensemble model predicts a better accuracy of 84.1% at the $20^{th}$ iteration than the LSTM model, which achieves the best accuracy among the baseline models.

Based on the experimental results, the eFedDNN model shows better performance when compared with any of the single classifiers of the federated-learning-based DNNs. Thus, our study is consistent with previous research which proposes that the performance of the ensemble model is sufficient to make better predictions of travel mode than individual single classifiers using the MTL Trajét dataset. The experimental research carried out over the last few years proves that when the outputs of various classifiers are combined, they can reduce generalization errors and deal with the high variance of individual classifiers. Our results suggest that the eFedDNN model not only ensures the protection of travellers' privacy but provides an accurate prediction of users' mode of travel.

*2) Performance Comparison of the Baseline FL Models :* The performance of the vanilla FL models of the deep

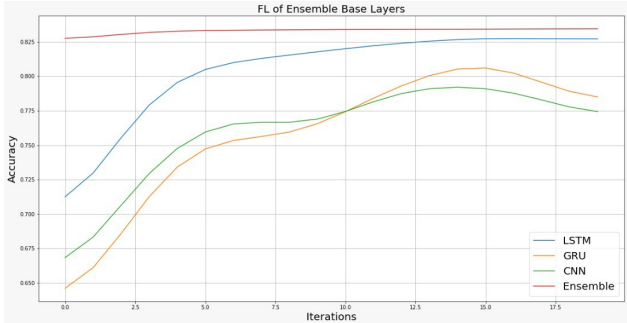

Fig. 3. The prediction accuracy of the baseline and ensemble models

TABLE I. Prediction Accuracies of FL Models

| Model Type | Iteration | Accuracy (%) |
|---|---|---|
| LSTM | 20 | 83.1 |
| GRU | 20 | 76.2 |
| 1D CNN | 20 | 75 |
| eFedDNN | 20 | 84.1 |
| CNN Ensemble(ML) | 20 | 80.0 |

neural networks is compared, as shown in Figure 3. In this study, we compare the prediction accuracy of the models trained in a similar FL architecture to corroborate the usage of the federated learning framework. The models stopped training at around the 20th iteration, and the number of communication rounds was set as 20 for federated learning. As shown in Figure 3 and Table 1, the prediction accuracy of the federated-trained models varies from each other.

Moreover, to determine the model with the best prediction accuracy, we trained all the models on the same input values with more length. LSTM outperforms the other baseline models, GRU and 1D CNN, by reducing the error of the mode of travel and improving prediction accuracy. As shown in Figure 3, with the test data, the LSTM can achieve the highest accuracy of 83.1% at the 20th iteration. At the same time, the GRU and 1D CNN obtain prediction accuracy of 76.2% and 75.0%, respectively, at the 20th iteration. Because it can distinguish between different modes more precisely than the other neural network models, the LSTM model has the highest prediction accuracy. The LSTM contains long-term and short-term memory that are utilized to fit different patterns, making it better suited to recognizing issues with continuous features. Hence, their travel modes change in various ways. For instance, in walk mode, the acceleration time is short, and the speed is stable and low; in bus and automobile mode, the acceleration time is longer, and the speed is higher.

*3) Performance Comparison of Proposed Model with Centralized Machine Learning models:* To demonstrate the superiority of the proposed model, we compare our results with the studies that have used the MTL Trajét dataset so that a fair comparison can be established. Table 2 shows the prediction accuracies of both the centralized and FL models. We compare the prediction accuracy of the proposed FL ensemble-based DNN model with the centralized ML ensemble-based CNN model by Yazdideh et. al for mode inference in a smartphone travel survey.

The authors employed different CNN models for travel mode inference in different configurations to develop the ensemble model in their study. The prediction accuracy of the centralized CNN model with only one convolutional layer is 72.5%, while the prediction accuracy of the federated-learning 1D CNN model is 75%. The final result of our experimentation shows significant improvement in the FL model compared with the centralized CNN model. Again, the prediction accuracy of the eFedDNN model is 84.1%, while the CNN ensemble-based model that employs average voting achieved a prediction accuracy of 85.0%. Specifically, the centralized ensemble CNN model has a prediction accuracy of 0.9% higher than eFedDNN model. The experimental results from both studies show that ensemble models outperform single base classifiers in identical configurations.

However, the superiority of the proposed ensemble model over the centralized ensemble model is based on the protection of users' private information in a real-world implementation. While the centralized ML model requires the upload of GPS data to a centralized location resulting in a breach of users' privacy, the ensemble FL framework considers users' privacy by ensuring that data are not shared in a centralized location. Hence, the proposed model does not only give an accurate prediction of travel mode but also ensures that the privacy of users is not violated.

TABLE II. Prediction Accuracies of FL and ML Models

| Architecture | Model Type | Accuracy (%) |
|---|---|---|
| FL models | eFedDNN | 84.1 |
|  | 1D CNN | 75 |
| ML Models | ensemble CNN | 85.0 |
|  | CNN | 72.5 |

V. Conclusion and Recommendation

In this paper, we propose an ensemble-based Federated Deep Neural Network (eFedDNN) architecture for travel mode inference. Three neural networks that include LSTM, GRU, and 1D CNN are used as base-learners, whereas MLP is used as the meta-learner for the stacking ensemble method. The model uses privacy mechanisms to train a global model in a distributed manner rather than allowing direct access to the user data. We constructed a set of Deep Neural Networks augmented by the stacking ensemble technique utilizing the FL framework to infer trip mode from GPS trajectories obtained from a large-scale smartphone travel survey. The FL approach involves aggregating the local model updates uploaded by all the locally trained models from the workers to the chief to build a global one for travel mode inference. The datasets of GPS trajectories are handled in such a way that they may be given as an input layer to a series of DNNs in privacy-preserving manner. Each trip has fixed segments with four channels that include relative distance, speed, acceleration, and jerk rate.

With the stacking ensemble method, we combined the results of the LSTM, GRU, and 1D CNN to obtain a better prediction accuracy than the baseline models. Our ensem-ble library included several models with various hyperparameter values and architectures. On the MTL trajét open-access dataset, we evaluated the performance of eFedDNN and compared it to the vanilla FL and non-federated learning methods. The findings demonstrate that the proposed ensemble technique outperforms the baseline models with better accuracy. Among the three neural networks of the vanilla neural network-based FL, the LSTM model has the best accuracy compared with GRU and 1D CNN.

In future work, we aim to test the proposed architecture under a split-learning setting in order to add a layer of defense against cyberattacks. We also aim to integrate Graph Convolutional Network into the ensemble-based FL architecture to test if we can better capture the spatial-temporal interdependence among trip mode data and increase prediction accuracy.

## Acknowledgment

This research is funded by Ryerson University, Ontario Early Researcher Award, and NSERC Canada Research Chair on Disruptive Transportation Technologies and Services.